\documentclass[10pt,twocolumn,letterpaper]{article}

\usepackage{iccv}
\usepackage{times}
\usepackage{epsfig}
\usepackage{graphicx}
\usepackage{amsmath}
\usepackage{amssymb}

\usepackage[pagebackref=true,breaklinks=true,letterpaper=true,colorlinks,bookmarks=false]{hyperref}

\iccvfinalcopy % *** Uncomment this line for the final submission

\def\iccvPaperID{****} % *** Enter the ICCV Paper ID here

\begin{document}

%%%%%%%%% TITLE
\title{AAN: Attributes-Aware Network for Temporal Action Detection}

\author{\large
Rui Dai\textsuperscript{1}\thanks{This work was performed while Rui was at Inria as a student.}, \hskip 1em
Srijan Das\textsuperscript{2},\hskip 1em
Michael S. Ryoo\textsuperscript{3},\hskip 1em
Fran\c cois Brémond\textsuperscript{1}%\hskip 1em
\\
\textsuperscript{1}Inria \hskip 1em
\textsuperscript{2}UNC Charlotte \hskip 1em
\textsuperscript{3}Stony Brook University \hskip 1em
\\
{\tt\small \textsuperscript{1}\{name.surname\}@inria.fr} %\hskip 1em 
% {\tt\small \textsuperscript{3}\{name.surname\}@stonybrook.edu} %\hskip 1em
}

\maketitle
\thispagestyle{empty}

\begin{abstract}
The challenge of long-term video understanding remains constrained by the efficient extraction of object semantics and the modelling of their relationships for downstream tasks. Although OpenAI's CLIP visual features exhibit discriminative properties for various vision tasks, particularly in object encoding, they are suboptimal for long-term video understanding. To address this issue, we present the \textbf{Attributes-Aware Network} (AAN), which consists of two key components: the Attributes Extractor and a Graph Reasoning block. These components facilitate the extraction of object-centric attributes and the modelling of their relationships within the video. By leveraging CLIP features, AAN outperforms state-of-the-art approaches on two popular action detection datasets: \textbf{Charades} and \textbf{Toyota Smarthome Untrimmed} datasets.
\end{abstract}

%-------------------------------------------------------------------------
\section{Introduction}
\label{sec:intro}
In video understanding, temporal action detection is one of the ultimate tasks that automatically detects human actions in videos along with classifying them. The deep learning revolution has been a huge driving force for the advancements in the video understanding domain.
%temporal action detection is a challenging and fundamental sub-field in computer vision that automatically detects human action in videos along with classifying them. % Among the sub-tasks in video understanding, how to detect fine-grained actions from long-untrimmed videos is an important and challenging problem. 
% 
Despite the progress of action recognition algorithms in trimmed videos~\cite{video_transformer_network,arnab2021vivit,i3d,das2021vpn++,C3D}, the majority of real-world videos are lengthy and untrimmed with dense regions of interest~\cite{charades,Dai_2022_PAMI}. In these regions (temporal intervals), most actions involve human interaction with objects, such as \textit{opening "fridge"}, \textit{taking "ham"}, and \textit{cutting "bread"}. These objects and their state changes are crucial attributes to understand human actions performed in videos. 
Thus, modelling these fine-grained object semantics for detecting actions is paramount in complex activities such as \textit{making breakfast} or \textit{furniture assembling}. 
% However, due to the limitation of the computation resources, the current methods can not leverage the objects in the video for untrimmed video in an effective manner. 

\begin{figure}[!t]
\centering
\includegraphics[width=.9\linewidth]{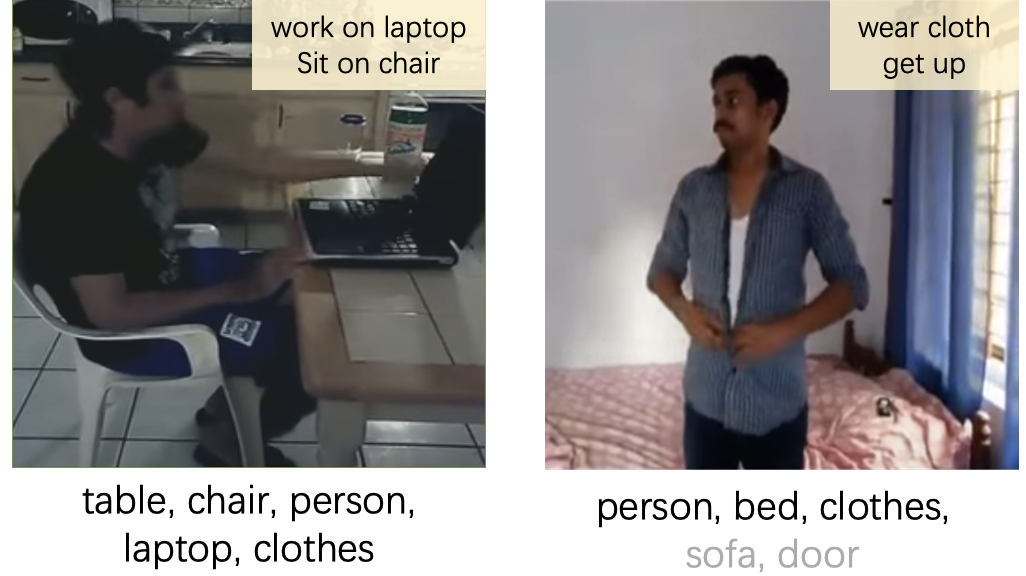}
\caption{CLIP image classification for the video frames. A list of daily living attributes (i.e., objects) is used as labels, i.e, the inputs to the CLIP Text model. We show the top 5 highest similarity attributes for two example frames (below) along with the action label (top right). We find that the CLIP Image features effectively preserve the action-relevant object semantics of an image. }
\label{fig:CLIPexample}
\end{figure}

Many successful action detection models have been developed to process untrimmed videos in two stages~\cite{TGM2, superevent, dai2021pdan, dai2022mstct}. In the initial stage, frame-level features are extracted from visual input using a 3D convolutional network~\cite{i3d} that has been pre-trained on extensive video datasets like Kinetics~\cite{kinetics}. The subsequent stage entails modeling temporal relationships among the frame features to detect activities. Nevertheless, these approaches do not explicitly encode object semantics. On one hand, limited by the absence of annotations of the relevant objects involved in action due to constrained labeling, the frame-level features may not preserve the object semantics relevant to the target actions in the first stage. On the other hand, the second stage is dedicated exclusively to temporal representation learning across the frame features.
Some methods~\cite{video-as-graph,ghosh2018stacked_stgcn} have attempted to enhance action understanding by employing object detectors and subsequently incorporating a reasoning module that operates on the extracted objects for action detection. While these frameworks are capable of efficiently extracting object semantics, the accuracy of action prediction is heavily dependent on the precision of object detection. Furthermore, the inclusion of an object detector introduces a trade-off. Object detectors are known for their large model complexity, often leading to increased computation costs during inference. Furthermore, methods using object detectors leverage region of interest (ROI) operations on intermediate 3D convolution features to optimize object detection~\cite{video-as-graph}. Nevertheless, this technique typically operates on a restricted temporal data sequence, potentially restricting the model’s capability to capture short-term relationships.

In pursuit of a dense understanding of scene, vision-language models, specifically OpenAI's CLIP~\cite{CLIP}, have demonstrated remarkable efficacy in pre-training image and text encoders for a variety of downstream tasks. Inspired by CLIP's success, numerous models have been pre-trained using large-scale open-vocabulary data comprising image-text pairs, resulting in a joint feature space for image and language~\cite{CLIP,shen2021much,xu2021vlm}. As the pre-training process is not limited to a predefined set of object labels, the visual representations obtained are aligned with a more extensive range of "language" semantics~\cite{lin2022frozen}. Due to this configuration, CLIP features retain a richer object semantics (refer to Fig.~\ref{fig:CLIPexample}). Consequently, in this paper, we explore the question: \textit{How can we leverage CLIP features for fine-grained action detection?}

%Recently, there are some methods based on Contrastive Language-Image Pre-training (CLIP) for bridging the gap between language  and image models~\cite{CLIP}. The CLIP-based models are pre-trained with large-scale open-vocabulary datasets with image–text pairs, which learned a joint feature space for image and language~\cite {CLIP,shen2021much,xu2021vlm}. Because the pre-training is not based on fixed labels, the learned visual representations are aligned with richer "language" semantics~\cite{lin2022frozen}. Thanks to this setting, the CLIP feature preserves richer semantics of the objects (see Fig.~\ref{fig:CLIPexample}). How can we utilize CLIP features for fine-grained action detection? 
%%%Srijan will continue from here
To this end, we propose the Attributes-Aware Network to address the challenge of multi-label action detection. This network is composed of two modules: Firstly, an attribute extractor that learns to extract attribute semantics from the frame-level features obtained from the CLIP. Different from the existing object-centric video recognition methods~\cite{video-as-graph,Guermalicprthorn,ghosh2018stacked_stgcn}, our method does not rely on the prior of the object detectors but leverages the joint visual and language space of CLIP features. The extracted attributes are relevant objects (e.g., knife) for an action (e.g., cutting) and the semantics of each attribute form nodes of a graph. 
Secondly, we introduce an attention-based graph reasoning module that models the inter-attribute relations and temporal relations of attributes within a video for frame-level action prediction.

To summarize, our contribution mainly lies in three folds. 
Firstly, we propose a module that extracts relevant object semantics of image frames using the CLIP. This module disentangles target information from the shared space between visual and semantic features. Secondly, we introduce an attention-based block to model the complex attribute relations within and across frames. Finally, our proposed method outperforms state-of-the-art approaches on two challenging multi-label action detection datasets. To our knowledge, we are the first to disentangle CLIP embeddings for the purpose of long-term video understanding.
%

% An effective real-world action understanding system should be able to detect multiple actions in long untrimmed videos. In this thesis, we focus mainly on temporal action detection in untrimmed videos, which aims at finding the action occurrences along time in the video.

\begin{figure*}[!t]
\includegraphics[width=\linewidth]{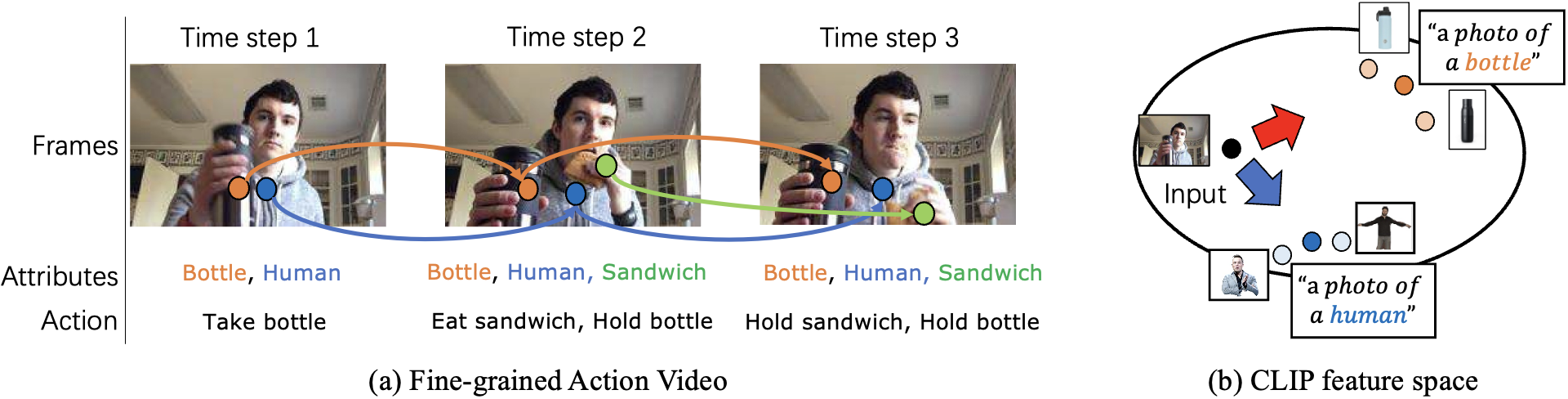}
\caption{On the left hand, there is an example of fine-grained actions in a video. Multiple actions can occur at the same time and actions can involve different objects (attributes). On the right hand, we show an example of the feature space of CLIP. In this space, the representation of the text sentence and image with the same semantics are close to each other. 
}
\label{fig:graphical_abstract}
\end{figure*}

\section{Related Work}
\label{sec:RW}
In this section, we survey existing approaches that model relations between different semantics for action understanding.

Lan et al.~\cite{lan2015action} propose a method that represents videos by a hierarchy of mid-level action elements in an unsupervised manner. Each action element corresponds to a spatio-temporal segment in the video and can represent actions at multiple spatio-temporal resolutions. 
Sigurdsson et al.~\cite{sigurdsson2017asynchronous} propose a fully-connected temporal CRF model for reasoning over the variant intent of videos, where the intent is defined as the clustering of similar activities (e.g., actions, objects) in a video. Although these approaches can structure the video using semantics, they do not explicitly learn the temporal structure nor are they learned in an end-to-end fashion.

In recent years, Wang et al.~\cite{video-as-graph} proposed video as graphs where the nodes are based on object proposals. 
Likewise, Ghosh et al.~\cite{ghosh2018stacked_stgcn} utilized the labels of the object bounding boxes to form fine-grained graphs of humans, scenes and objects for each image frame. 
Guermal et al.~\cite{Guermalicprthorn} extract object-specific feature descriptors for each object using an object detector and learn action correlations using an attention mechanism. While these methods better characterize complex object-based actions in videos, they rely on object detectors pre-trained on a predefined set of object categories, which limits their ability to handle unseen objects and increases the computation complexity at both training and inference time. %Furthermore, these frameworks root in video-level multi-label classification as the mechanism is designed for summarizing the video content. 

More recently, several methods~\cite{wang2021actionclip, Luo2021CLIP4Clip, bain2022cliphitchhikers, kahatapitiya2023victr} have used CLIP features for video understanding. However, these methods are designed to handle short temporal videos, and the challenge of handling actions over a long range of time for solving the task of action detection still persists. 
Towards long-term video understanding, Tirupattur et al.~\cite{MLAD} introduced MLAD that can explore the action-temporal relations with a set of self-attention layers: an inter-class attention map for every time step and an inter-time attention map for every action class. 
Similarly, Dai~\cite{dai2021ctrn} propose CTRN that can model the interaction relations via graph neural networks. However, both methods overlook object attributes, limiting their performance over object-dominated actions. 
In contrast, this paper proposes a method that learns relations among attributes (i.e., objects) extracted from CLIP features. To the best of our knowledge, this is the first method that leverages CLIP features for long-term video understanding while implicitly modelling object attributes. 

Besides, model relations between different semantics. There are also some works using pure temporal modelling for temporal action detection~\cite{TGM2,dai2022mstct, dai2021pdan}. TGM~\cite{TGM2} is a temporal filter based on Gaussian distributions, which enables the learning of longer temporal structures with a limited number of parameters. PDAN~\cite{dai2021pdan} is a temporal convolutional network, with temporal kernels which are adaptive to the input data. MSTCT~\cite{dai2022mstct} uses convolutions in a token-based architecture to promote multiple temporal scales of tokens, and to blend neighbouring tokens imposing a temporal consistency with ease. Different from the above methods, besides the temporal modelling, our proposed method further models the object semantic relation for a better understanding of the video content. 

\section{Proposed Method}
\label{sec:PM}

\begin{figure*}[!ht]
\centering
\includegraphics[width=.7\linewidth]{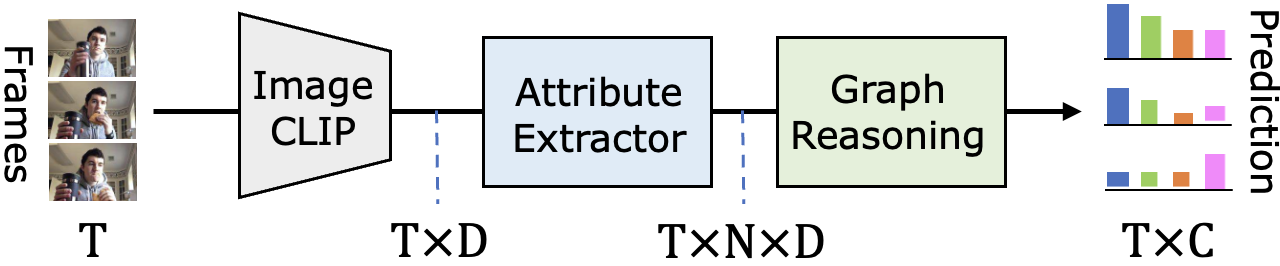}
\caption{Overall framework for the proposed Attribute-Aware Network. This network is composed of 3 main components: CLIP encoder, attribute extractor and graph reasoning block for graph classification. }
\label{fig:overall}
\end{figure*}

In this section, we present our \textbf{Attributes-Aware Network (AAN)}, which leverages CLIP features for the task of action detection. AAN is composed of three main components: a frozen CLIP encoder, an attributes extractor and a graph reasoning block (see Fig.~\ref{fig:overall}). The attributes extractor extracts the attribute semantics from the frame-level CLIP features. 
Conversely, the graph reasoning block models the attribute relation and performs graph classification. These two components of AAN are trained end-to-end to optimize the attribute representation for the action detection task. We elaborate on these components in the following. 

\subsection{Attributes Extraction}
\begin{figure*}[!t]
\centering
\includegraphics[width=.7\linewidth]{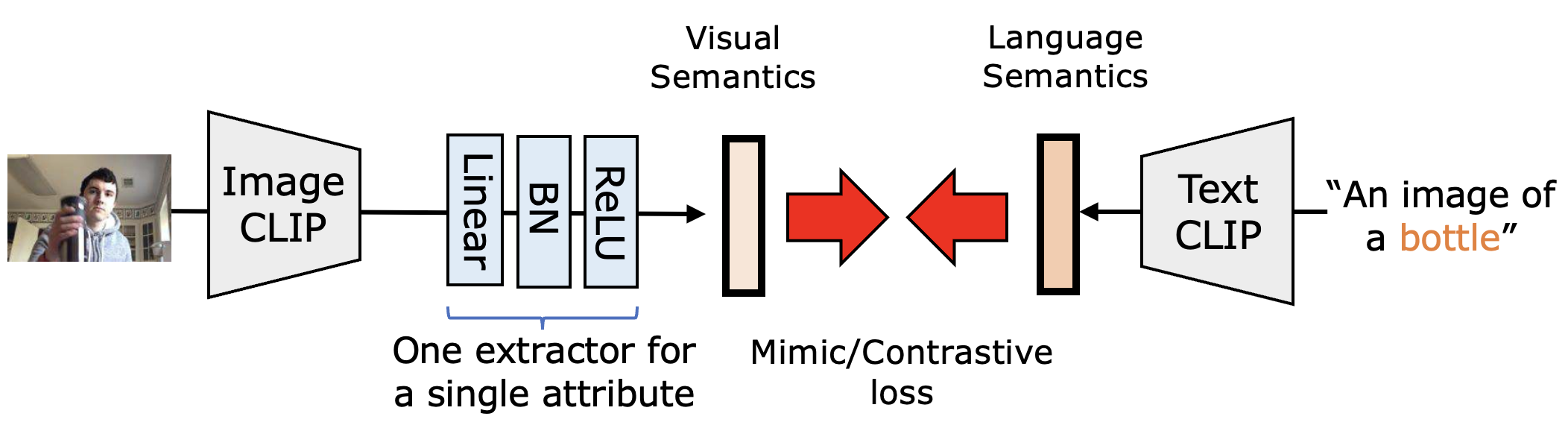}
\caption{Attributes Extractor. For each attribute, it has a specific extractor to obtain its semantic representation from the input frame. The extractor for the same attribute is shared across the frames.}
\label{fig:AE}
\end{figure*}

The attributes extractor is used to extract the attributes from the frames. These attributes are object semantics encoded within the frame representations. 
Different from the previous works~\cite{video-as-graph,ghosh2018stacked_stgcn} that extract the object semantics via object detector from the 3D convolutional feature map, in this work, the attributes extraction is based on the frame level representation obtained from CLIP encoder.

%frozen frame-level CLIP features and not relying on the object detector. 

%CLIP~\cite{CLIP} is composed of a pair of encoders: Image encoder and Text encoder, which is trained on . In the training phase, the method does not rely on fixed labels but on open-world sentences. The image-sentence pairs are fed into the two encoders. The learning process is based on the minimization of the distance between the image feature and its description (i.e., sentence) feature in a contrastive manner. 
%
CLIP~\cite{CLIP} is trained using a vast collection of image-text pairs in a contrastive fashion. Owing to the CLIP pre-training process, language and image features are projected into a shared embedding space, as depicted in Fig.\ref{fig:graphical_abstract}(b).
This configuration enables the CLIP Image model to compute more generalized visual features, as demonstrated in \cite{shi2022proposalclip,wang2021actionclip,gu2021open} for a range of vision tasks.
As illustrated in Fig.~\ref{fig:CLIPexample}, CLIP Image features effectively retain the action-relevant object semantics within an image. However, CLIP Image features consist of a blend of various object representations. This raises the question: \textit{how can we disentangle object semantics from the CLIP Image feature for long-term video understanding}?

% As shown in Fig.~\ref{fig:graphical_abstract}(b), thanks to the CLIP pre-training, the language and image features are projected in a joint space~\cite{CLIP}. 
% As mentioned earlier, the CLIP model projects the language and image features in the same space. 

In the joint embedding space of CLIP, when given a text prompt describing a specific object, such as "a photo of a \textit{bottle}", the feature embedded by the Text CLIP encoder conveys the pure semantics of "bottle" in this space. This feature is closely related to images of "bottle" embedded by the Image CLIP model (refer to Fig.~\ref{fig:graphical_abstract}(b)). These prompts can be considered as anchors within the joint space, allowing the disentanglement of particular object semantics from the holistic image representation. Therefore, in this work, the semantics for a specific object are computed by minimizing the Euclidean distance between the visual frame feature and the text anchor representation in the shared semantic space. 

In this work, our focus is on indoor environments. Initially, we pre-define ${N}$ attributes associated with daily living actions. Since action labels typically consist of a noun and a verb, the attributes should encompass all nouns in the datasets. Next, we generate prompts based on these attributes and employ the CLIP Text encoder to extract text anchor features in the joint embedding space, represented by $\mathcal{T}^n$, where $n\in N$.

In the visual aspect, frame-level features are obtained using the CLIP Image encoder (refer to Fig.~\ref{fig:AE}). These extracted features are stacked along the temporal axis (i.e., frames) to create a $T \times D_0$ video representation. This video representation is then fed to the attributes extractor, which comprises ${N}$ filters. Each filter corresponds to a specific attribute and includes a linear layer, batch normalization, and ReLU activation, as follows:

\begin{equation}
I^n _t = ReLU(BN(W^n F_t))
\end{equation}
Here, $F_t$ represents the CLIP feature of the frame at time step $t$, and $W^n$ denotes the linear layer for attribute $n$. ReLU and BN refer to ReLU non-linear activation and batch normalization, respectively. Note that each filter signifies an object semantics in the latent space.
We minimize the L2 distance~\cite{allen1971mean} between the output feature ($I^n _t$) of the attributes extractor and its corresponding text anchor feature ($\mathcal{T}^n$). This minimization objective encourages the extractor to extract the object-specific attributes in an image. The formulation of this objective is as:
\begin{equation}
     \mathcal{L}_{attributes} = \frac{1}{{N}{T}} \sum_{t \in T} \sum_{n \in {N}}|| {I^n _t} - {\mathcal{T}^n} || ^2
\end{equation}
Therefore, the linear layers situated above the frozen CLIP visual encoder, optimized with the aforementioned loss, facilitate the extraction of relevant attributes (object-related information) pertaining to the scene for video understanding.
%Note that there is recent work \cite{sung2022vl} that utilizes MLP in the CLIP model to transfer knowledge to different question-and-answer tasks. Different from the previous work, here we utilize the linear layers to extract specific attribute semantics from the CLIP Image feature to understand the scene for action detection better. 
% 
Note that, it is possible to utilize both MSE loss and contrastive loss~\cite{Dai_2021_ICCV,moco} for semantic extraction. In our work, we have explored the use of contrastive loss but found that L2 loss performs better for our specific task. This observation aligns with our intuition, which suggests that contrastive learning is more suitable for large-scale pre-training rather than fine-tuning downstream tasks. Balancing embedding learning with downstream task classification can be challenging, making MSE loss a more effective choice for our purposes. 

\subsection{Graph Reasoning Block}

Following the attribute extraction, each frame can be represented as a graph, where the extracted attributes function as the graph nodes. The graph edges are initialized based on the statistics of attribute occurrences in the training distribution.
An action label consists of a noun and a verb. The occurrence of attributes is determined by the co-occurring probability of the nouns. $G_{ij}$ represents the concurrent instances for attribute classes $n_i$ and $n_j$. The conditional probability matrix $P_{ij} = P(n_j | n_i)$ is then calculated as $P_{ij} = G_{ij}/G_i$, where $G_i$ denotes the frequency of $n_i$ in the training distribution, and $P_{ij}\in\mathbb{R}^{N\times N}$ indicates the probability of class $n_j$ given the simultaneous occurrence of $n_i$. The computed $P$ represents the initial graph edges. % We also added a learnable weight mask on top of the edge's adjacency matrix. This weight mask is optimized with the final objective, thus it can learn the attribute-attribute relation related to the action classification.
All graphs (i.e., frames) within the video share the same graph initialization settings.
The action prediction for each frame can be interpreted as a graph classification task.

\begin{figure*}
\centering
\includegraphics[width=.7\linewidth]{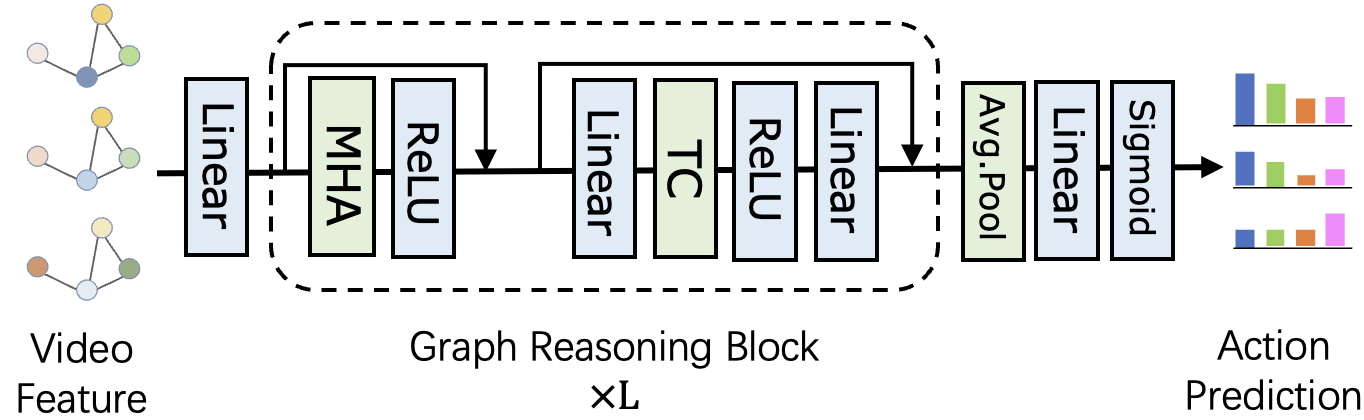}
\caption{Graph Classification Module.
}
\label{fig:GR}
\end{figure*}

In this work, our graph classification mechanism is close to Self-Attention Graph Pooling (SAGPool)~\cite{lee2019self}. However, in contrast to~\cite{lee2019self}, we perform graph reasoning on a series of frames, which includes modelling both attribute-attribute and attribute-temporal relations. %Moreover, this module should benefit from both shared attributes relation and the relation of the specific attributes across frames. 
As shown in Fig.~\ref{fig:GR}, this block comprises $L$ attention-based graph reasoning layers followed by a pooling layer to aggregate the graph-level information. Each graph reasoning block contains a multi-head graph convolution layer, a Temporal Convolution layer, and a linear layer with non-linear activations and residual links. 

The input feature is first fed into a linear layer (i.e., bottleneck layer) to squeeze the feature dimension from $D_0$ to $D_1$. Then the features are fed to the graph reasoning blocks. In each block, self-attention-based graph convolution is applied on each frame. The graph convolution is used for modeling the relationship among the attributes at each frame. 
% 
% \textcolor{red}{
In practice, for block $i$ at frame $t$, the input features is $X^i _t \in \mathbb{R}^{D_1 \times 1 \times N}$. Then we utilize a self-attention module to learn the inter-attributes relation. Moreover, the computed relation is then summed up with the initial graph edge based on the co-occurring attributes probability matrix $P$. 
% Then, an $N \times N$ attention matrix is summed up with the initial graph edge $P$. 
The obtained graph adjacency matrix is formulated as follows:
\begin{equation}
    A^i _t = softmax((W^i _1 X^i _t)^\top  W^i _2 X^i _t) + P
\end{equation}
where $W^i _1$ and  $W^i _2 \in \mathbb{R}^{D_1 \times D_1}$ are the weights of two linear layers. Each value in $A_t$ can be seen as a composite edge between two vertices. 
$P$ represents the global co-occurrence statistics from the training set and the self-attention mask represents the relation that is adaptive to different frames. % }
% The learned graph is shared across different frames but unique for different layers and videos. 
% This design choice can capture the inter-class dependencies in a video and makes CGCN scalable across different temporal scales. 
Finally, the graph convolutional operation is performed using the formulation from~\cite{kipf2016semi}: 
\begin{equation}
   X'^{i} _t = ReLU(A^i _t X^i_t W^i _3) + X^i_t
\end{equation}
where $W^i _3 \in \mathbb{R}^{D_1 \times D_1}$ is the weight of the linear layer. 
The output feature of the graph convolution is then fed into a linear layer, followed by a temporal convolution. The linear layer is employed for channel mixing prior to applying the temporal convolution. Temporal convolution is performed for the same node across multiple frames to model temporal information. This is followed by a ReLu activation and an additional linear layer. These operations can be expressed as:

\begin{equation}
X^{i+1} = W^i _5(ReLU(TC(W^i _4 X'^{i}))) + X'^{i}
\end{equation}
where $W^i _4$ and $W^i _5 \in \mathbb{R}^{D_1 \times D_1}$ are the weights of two linear layers, while $TC$ denotes the temporal convolution operation. The output $X^{i+1}$ is input to the subsequent graph reasoning blocks.  
%
% We can pre-defined the link of the graph
% initialized learnable weights. 
% learn the initial link from the training set. 

\iffalse
\begin{equation}
    \mathcal{L}_{att}= - \frac{1}{n} \sum^{n}_{i=1} (a^j_i(1-a^j_i)\beta^j_k log(1-P_{att}(x_i,a^j_i)))
\end{equation}
\fi
In the final step, graph classification is carried out by classifying the aggregated graph representation for each frame. The graph aggregation is performed using an average pooling layer, while the classification of the graph employs a linear layer with sigmoid activation. The Binary Cross Entropy (BCE) loss, $\mathcal{L}_{action}$, is computed in comparison to the ground-truth labels for multi-label action classification of each frame, as demonstrated in~\cite{chen2019learning}.
Consequently, the total loss for training AAN is given by:

\begin{equation}
\mathcal{L}_{total}= \mathcal{L}_{attributes} + \mathcal{L}_{action}
\end{equation}

% \newpage

\section{Experiment}
\label{sec:Exp}

\subsection{Dataset}
In this work, we evaluate our method on two challenging action detection datasets which involve actions with fine-grained object details. \textbf{Charades}~\cite{charades} is a
large untrimmed dataset with 9848 videos of daily living actions. The dataset contains 157 action classes with more than 30 objects shared across multiple action classes. We utilize the "action localization" setting for this dataset, which aims at detecting actions for different frames~\cite{sigurdsson2017asynchronous}. We also evaluate our method on \textbf{Toyota Smarthome Untrimmed (TSU)}~\cite{Dai_2022_PAMI, smarthome_iccv} (Cross-subject protocol). Similar to the Charades, TSU is recorded in an indoor environment. There are up to 5 actions that can occur at the same time in a given frame. Different from Charades, the TSU involves long-term videos and composite activities. 
For evaluation, we compute the per-frame mAP by default on these two datasets following~\cite{multi-thumos}.

\subsection{Implementation details}
In the proposed network, we employ ViT/14~\cite{dosovitskiy2020image} based CLIP visual encoder. The CLIP encoding feature size $D_0$ is 768. For light-weighting the network, we then map the $D_0$ to intermediate channel size $D_1$, which is 256. We set the number of attributes $N$ to 38 to fit the requirement of general daily living video understanding. 
There are $L=5$ graph reasoning blocks used, and the kernel size for the temporal convolution within the graph reasoning block is 3. The number of heads for the multi-head attention is set to 4.
%$C$ corresponds to the action classes of the datasets. 
% 
AAN is trained using two RTX 6000 GPUs with a batch size of 32. The Adam optimizer~\cite{adam_optimizer} is utilized with an initial learning rate of 0.0001, which is scaled by a factor of 0.5 with a patience of 8 epochs.

\noindent\textbf{Prompt}: In our work, we perform feature extraction on a per-frame setting. As a result, we utilize a standard image prompt for image classification. Specifically, during the training phase, we define a list of prompts including (1) "\textit{A photo of} \textbf{xx}", (2) "\textit{There is a} \textbf{xx}", (3) "\textit{An image of} \textbf{xx}", and (4) "\textit{A photo with a} \textbf{xx}", where \textbf{xx} represents the object label. To enhance the robustness of attribute representation, we randomly select one of these prompts for each video during training. During inference, we use prompt (1) to extract attributes.

\noindent\textbf{Attributes}: The predefined attributes in our study are derived from the object and action labels (e.g.~"\textit{book}" in "\textit{reading book}") provided by the Charades and TSU datasets. Additionally, both datasets include a list of objects present in their respective datasets, and we leverage this information to compile our attribute list.

\subsection{Comparison to the State-of-the-Art}
In this section, we compare AAN with state-of-the-art methods on two large indoor datasets, Charades~\cite{charades} and TSU~\cite{Dai_2022_PAMI}, as shown in Table~\ref{table:comparaison_SOTA}. Note that, similar to our method, we compared only the RGB only result. Both datasets feature complex actions with varying object interactions. We compare our approach with leading methods for these two datasets, including techniques utilizing TCN~\cite{dai2021pdan}, Transformer~\cite{MLAD, ttm}, graph convolution~\cite{dai2021ctrn}, and ConvTransformer~\cite{dai2022mstct}. We observe that our method significantly outperforms state-of-the-art approaches (e.g., +3.2\% on Charades and +7.4\% on TSU compared to MS-TCT~\cite{dai2022mstct}). This marks the first time a method achieves 30\% in the localization task on the Charades dataset and 40\% on the TSU dataset. In our model analysis, we demonstrate that this substantial improvement in action detection performance is not solely due to the use of the CLIP visual encoder in comparison to I3D~\cite{i3d} or X3D~\cite{x3d} encoders employed in state-of-the-art methods. Rather, it is attributed to each component of AAN, which plays a crucial role in leveraging the CLIP features for the action detection task.

Additionally, we assess the performance of our method using the action dependency metrics~\cite{MLAD} on the Charades dataset. As depicted in Table~\ref{table:dependency_comparaison_SOTA}, our method surpasses MLAD and MS-TCT across all metrics for both co-occurring action detection ($\tau$ = 0) and distant action detection ($\tau$ = 40). This demonstrates the robustness of our proposed approach.

\begin{table}[t!]%\tabcolsep=9pt
\centering
\caption{Comparison with the state-of-the-art methods. % Similar to MLAD, both RGB and Optical flow are used for the evaluation. $P_{AC}$ - Action-Conditional Precision, $R_{AC}$ - Action-Conditional Recall, $F1_{AC}$ - Action-Conditional F1-Score, $mAP_{AC}$ - Action-Conditional Mean Average Precision. $\tau$ indicates the temporal window size. $\tau = 0$ corresponds to the actions occurring at the same time.
}
{
\begin{tabular}{l|cc}
\hline
Eval in per-frame mAP (\%)            & Charades & TSU                  \\\hline
R-C3D~\cite{RC3d}           & 12.7                    & 8.7                  \\
Super-event~\cite{superevent}   & 18.6                       & 17.2                 \\
TGM~\cite{TGM2}       & 20.6                      & 26.7                 \\
PDAN~\cite{dai2021pdan}      & 23.7                      & 32.7                 \\
Coarse-Fine~\cite{kahatapitiya2021coarse} & 25.1                        & -                    \\
MLAD~\cite{MLAD}        & 18.4                        & -                    \\
CTRN~\cite{dai2021ctrn} & 25.3 & 33.5 \\
MS-TCT~\cite{dai2022mstct}   & {25.4}        &{33.7}\\
Coarse-fine + SSDet~\cite{kahatapitiya2023ssdet}   &   26.9    & - \\ 
ViVit-L + TTM~\cite{ttm}& 28.8 & - \\ 
\hline\hline
\textbf {Attribute-Aware Network}  & \textbf{32.0}           & \textbf{41.3} \\\hline
\end{tabular}
}
\label{table:comparaison_SOTA}
 % \vspace{-0.1in}
\end{table}

\subsection{Further Analysis}

\subsubsection{Ablation}
\noindent\textbf{Model Analysis.}
In this section, we examine the effectiveness of each component in AAN on the Charades dataset. Table~\ref{Tab:Ablation1} demonstrates our analysis of the necessity of the Attribute Extractor and the Graph Reasoning block. Note that the graph reasoning block can not operate without the attribute extractor. 
As seen in the table, merely extracting object attributes from the CLIP feature does not significantly enhance performance (+1.4\%). However, performing reasoning across these extracted attributes leads to better modelling of complex videos (+14.0\%).

Furthermore, we assess the importance of components within the Graph Classification block: Multi-Head Attention (MHA) and Temporal Convolution. These results are computed based on the presence of the Attribute Extractor.
Table~\ref{Tab:Ablation_gr} reveals that adding MHA or Temporal Convolution both improve performance compared to the vanilla Attribute Extractor (+7.0\% and +8.3\%, respectively). By incorporating all components, AAN achieves a substantial improvement.

\begin{table}[t]
\caption{Ablation on the Proposed Modules.}\vspace{2mm}
\centering
\scalebox{0.95}{
\begin{tabular}{cc|c}
\hline
Attribute Extractor & Graph Reasoning Blocks & \multicolumn{1}{c}{mAP (\%)} \\\hline
& & 18.1 \\%
$\checkmark$                   &                & 19.4                    \\% \hline
$\checkmark$                     & $\checkmark$                & 32.0                 \\\hline    
\end{tabular}}
\label{Tab:Ablation1}
\end{table}

\begin{table}[t]
\caption{Ablation inside graph classification module.}\vspace{2mm}
\centering
\scalebox{0.95}{
\begin{tabular}{cc|c}
\hline
Multi-Head Attention& Temporal Conv. & \multicolumn{1}{c}{mAP (\%)} \\\hline
$\checkmark$ &  & 25.1 \\%
&$\checkmark$                              & 26.4                   \\% \hline
$\checkmark$                     & $\checkmark$                & 32.0                 \\\hline    
\end{tabular}}
\label{Tab:Ablation_gr}
\end{table}

\begin{table}[t]
\centering
\caption{Ablation on the visual backbone. }\vspace{2mm}
\scalebox{0.95}{
\begin{tabular}{l|c|c|c|c}
\hline
    & MLP & PDAN~\cite{dai2021pdan} & MS-TCT~\cite{dai2022mstct} & Ours \\\hline
I3D~\cite{i3d} & 15.6    & 23.7     & 25.4       & -     \\\hline
ViT~\cite{CLIP} & 19.0    & 26.0     & 29.7       & 32.0     \\\hline
\end{tabular}}
\label{tab:backbone}
\end{table}

\begin{table*}[t!]%\tabcolsep=9pt
\centering
\caption{Evaluation on the Charades dataset using the action-conditional metrics~\cite{MLAD}. % Similar to MLAD, both RGB and Optical flow are used for the evaluation. $P_{AC}$ - Action-Conditional Precision, $R_{AC}$ - Action-Conditional Recall, $F1_{AC}$ - Action-Conditional F1-Score, $mAP_{AC}$ - Action-Conditional Mean Average Precision. $\tau$ indicates the temporal window size. $\tau = 0$ corresponds to the actions occurring at the same time.
}
\label{tab_cooccuring}
{
\begin{tabular}{l|ccc|ccc|ccc}
\hline
     & \multicolumn{3}{|c|}{$\tau$ = 0} & \multicolumn{3}{|c|}{$\tau$ = 20} & \multicolumn{3}{|c}{$\tau$ = 40} \\\hline
     & $P_{AC}$   & $F1_{AC}$   & $mAP_{AC}$  & $P_{AC}$   & $F1_{AC}$   & $mAP_{AC}$   & $P_{AC}$    & $F1_{AC}$   & $mAP_{AC}$   \\\hline
I3D~\cite{i3d}  &14.3          &2.1      &15.2      &12.7         &2.9      &21.4       &14.9         &3.1      &20.3       \\
% CF~\cite{}   &10.3     &1.0     &1.6      &15.8      &9.0     &1.5     &2.2      &22.2       &10.7     &1.6     &2.4      &21.0       \\
MLAD~\cite{MLAD} &19.3       &8.9      &28.9      &18.9         &10.5      &35.7       &19.6         &10.8      &34.8       \\
{MS-TCT}~\cite{dai2022mstct} &{26.3}       &{19.5}     &{30.7}      &{27.6}         &{22.1}      &{37.6}       &{27.9}        & 22.1      &{36.4}      \\\hline
\textbf{Attribute-Aware Network} &\textbf{31.4}&\textbf{20.4}&\textbf{35.4}&\textbf{30.4}&\textbf{22.3}&\textbf{41.8}&\textbf{32.5}&\textbf{22.2}& \textbf{40.8}     \\\hline
\end{tabular}
}
\label{table:dependency_comparaison_SOTA}
 % \vspace{-0.1in}
\end{table*}

\noindent\textbf{Backbone.}
% \textcolor{red}{
As our method is built on top of the CLIP backbone (i.e., pre-trained ViT model~\cite{dosovitskiy2020image}) rather than the conventional I3D model~\cite{i3d}. We thus further analyse if the performance boost is principal because of changing the backbone network. As shown in Table~\ref{tab:backbone}, we first evaluate the state-of-the-art method PDAN~\cite{dai2021pdan} and MS-TCT~\cite{dai2022mstct} with the pre-trained ViT backbone (i.e., CLIP Image model~\cite{CLIP}). We find that while using the same backbone as our method, the performance of PDAN and MS-TCT can be improved. However, as PDAN and MS-TCT do not have a specific design for leveraging the object-related feature, our method can still perform better (+2.3\%). 

We further compared our method and MS-TCT in terms of per-action class precision on Charades. We observe that for 22.9\% action classes, our method outperforms MS-TCT for more than 5\%. The top-5 actions that outperform MS-TCT are: Closing a window (+37.2\%), Sitting on a chair (+26.4\%), Taking a broom (+23.6\%), Closing a fridge (+22.0\%), Putting a laptop (+18.9\%). All the classes are relevant to objects.
% }

\begin{figure}[ht]
\centering
\includegraphics[width=.9\linewidth]{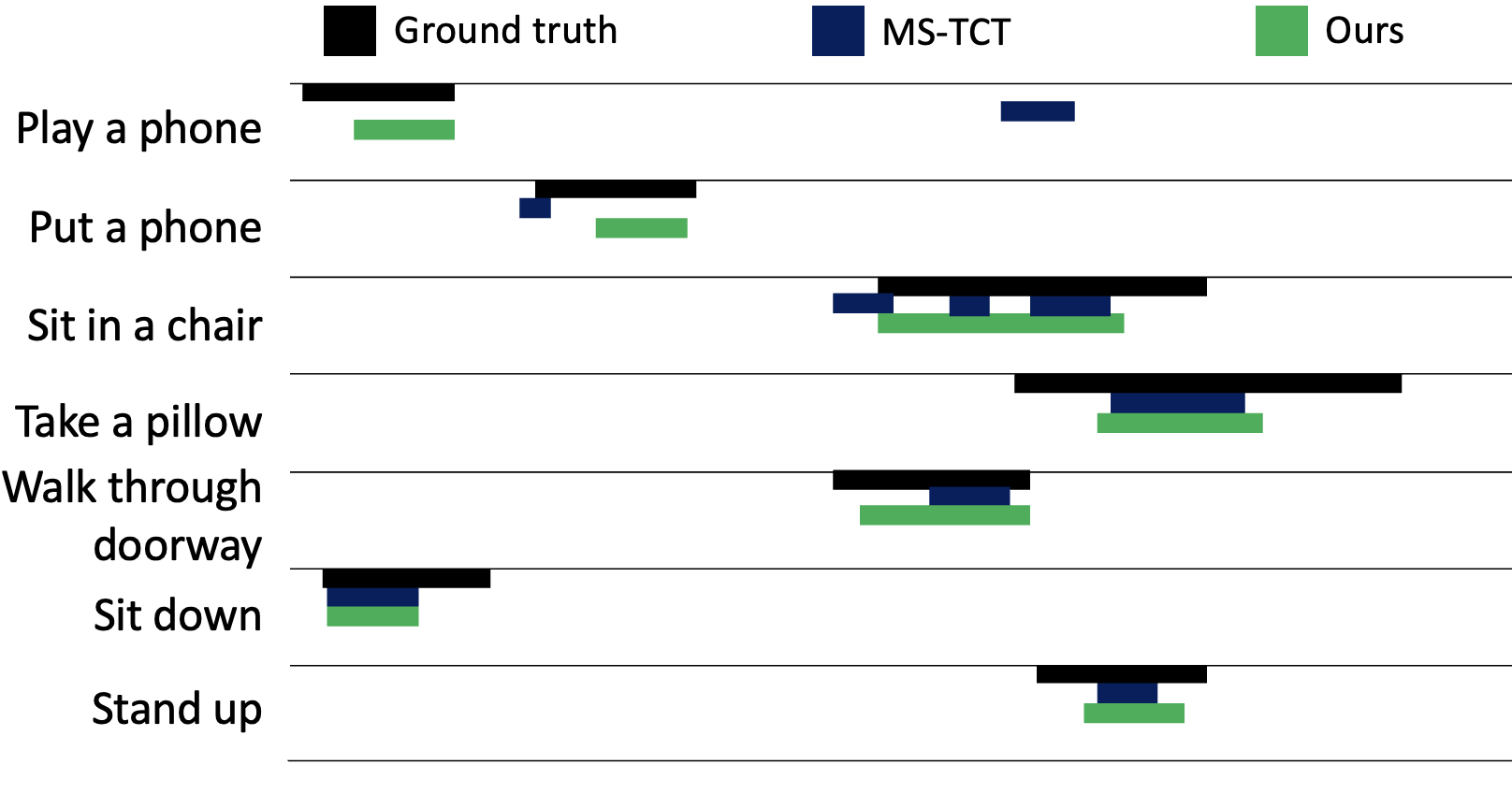}
\caption{Visualization of action detection.}\label{fig:vis}
\end{figure}

\subsubsection{Qualitative analysis}
As shown in Fig.~\ref{fig:vis}, we visualize the predictions of our method and the state-of-the-art method MS-TCT on a sample video from the Charades dataset. Both methods employ the same backbone network. We observe that, in comparison to MS-TCT, our method is more proficient in predicting object-related actions, such as \textit{play a phone}, \textit{put a phone}, and \textit{sit in a chair}.

\section{Conclusion}
In this paper, we introduced the Attributes-Aware Network (AAN), which utilizes CLIP features for action detection tasks. AAN comprises two essential components: the Attributes Extractor and the Graph Reasoning block, which are vital for learning object semantics and modelling their relationships in videos. AAN surpasses previous state-of-the-art methods on two widely-used Activities of Daily Living datasets, establishing a new benchmark. Future research will focus on rethinking various vision tasks using CLIP features and AAN-style frameworks. \\

\noindent\textbf{Acknowledgement:} This work has been supported by the French government, through the 3IA Cote d’Azur Investments in the Future project managed by the National Research Agency (ANR) with the reference number ANR-19-P3IA-0002. This work was also supported in part by the National Science Foundation (IIS-2245652). The authors are also grateful to the OPAL infrastructure from Université Côte d’Azur for providing resources and support.

%\newpage
{\small
\bibliographystyle{ieee_fullname}
\bibliography{egbib}
}
\end{document}